\PassOptionsToPackage{table,dvipsnames}{xcolor}
\documentclass[a4paper,twoside]{article}

\usepackage{comment}
\usepackage{adjustbox}
\usepackage{tikz}
\usetikzlibrary{positioning}
\usetikzlibrary{calc}
\usepackage{amsmath}
\usepackage{siunitx}
\usepackage[table, dvipsnames]{xcolor}
\usepackage{colortbl}
\usepackage{todonotes}
\usepackage{multirow}
\usepackage{makecell}
\usepackage{array}
\usepackage{tabularx}

\definecolor{lightgray}{gray}{0.85}

\usepackage{epsfig}
\usepackage{subcaption}
\usepackage{calc}
\usepackage{amssymb}
\usepackage{amstext}
\usepackage{amsmath}
\usepackage{amsthm}
\usepackage{multicol}
\usepackage{pslatex}
\usepackage{apalike}
\usepackage{algorithm2e}
\usepackage[bottom]{footmisc}
\usepackage{SCITEPRESS}     
\usepackage{hyperref}

\usepackage{ifthen} 
\newboolean{preprint} 
\setboolean{preprint}{true} 
\ifthenelse{\boolean{preprint}}%
    {
        \newcommand{\PREPRINTYEAR}{2025}
        \newcommand{\PUBLISHEDINSHORT}{ICINCO 2025}
        \newcommand{\PUBLISHEDIN}{22nd International Conference on Informatics in Control, Automation and Robotics  (\PUBLISHEDINSHORT)}
        \newcommand{\DOI}{10.5220/0013789200003982} 
        
        \usepackage[placement=top,vshift=-7]{background}
        \SetBgScale{1.0}
        \SetBgContents{\PUBLISHEDINSHORT. PREPRINT VERSION - DO NOT DISTRIBUTE. \href{https://doi.org/\DOI}{DOI \DOI}}
        \SetBgColor{black}
        \SetBgAngle{0}
        \SetBgOpacity{1.0}
    }

\begin{document}
\ifthenelse{\boolean{preprint}}%
    {
        \thispagestyle{empty}
        \onecolumn
        {
          \topskip0pt
          \vspace*{\fill}
          \centering
          \LARGE{%
            \copyright{} \PREPRINTYEAR~\PUBLISHEDIN\\\vspace{1cm}
            Personal use of this material is permitted.
            Permission from \PUBLISHEDIN \footnote{\url{https://icinco.scitevents.org/}}~must be obtained for all other uses, in any current or future media, including reprinting or republishing this material for advertising or promotional purposes, creating new collective works, for resale or redistribution to servers or lists, or reuse of any copyrighted component of this work in other works.}
            \vspace*{\fill}
        }
        \NoBgThispage
        \twocolumn          	
        \BgThispage
    }

\title{Towards Fully Onboard State Estimation and Trajectory Tracking for UAVs with Suspended Payloads}

\author{\authorname{Martin Jiroušek\sup{1}\orcidAuthor{0009-0003-4552-9995}, Tomáš Báča\sup{1}\orcidAuthor{0000-0001-9649-8277} and Martin Saska\sup{1}\orcidAuthor{0000-0001-7106-3816}}
\affiliation{\sup{1}Department of Cybernetics, Faculty of Electrical Engineering, Czech Technical University in Prague}
\email{martin.jirousek@fel.cvut.cz}
}


\abstract{
This paper addresses the problem of tracking the position of a cable-suspended payload carried by an unmanned aerial vehicle, with a focus on real-world deployment and minimal hardware requirements. In contrast to many existing approaches that rely on motion-capture systems, additional onboard cameras, or instrumented payloads, we propose a framework that uses only standard onboard sensors—specifically, real-time kinematic global navigation satellite system measurements and data from the onboard inertial measurement unit—to estimate and control the payload’s position. The system models the full coupled dynamics of the aerial vehicle and payload, and integrates a linear Kalman filter for state estimation, a model predictive contouring control planner, and an incremental model predictive controller. The control architecture is designed to remain effective despite sensing limitations and estimation uncertainty. Extensive simulations demonstrate that the proposed system achieves performance comparable to control based on ground-truth measurements, with only minor degradation ($< 6 \%$). The system also shows strong robustness to variations in payload parameters. \href{https://mrs.fel.cvut.cz/papers/uav-with-cable-suspended-payload}{Field experiments} further validate the framework, confirming its practical applicability and reliable performance in outdoor environments using only off-the-shelf aerial vehicle hardware.
}

\onecolumn \maketitle \normalsize \setcounter{footnote}{0} \vfill

\section{\uppercase{Introduction}}
\label{sec:introduction}

Unmanned aerial vehicles (UAVs) are playing an increasingly vital role across a wide range of applications, from aerial mapping~\cite{colomina_unmanned_2014}, infrastructure inspection~\cite{sikora_towards_2023}, and precision agriculture~\cite{gode_multi-stage_2024}, to last-mile delivery~\cite{murray_flying_2015} and environmental monitoring~\cite{c_path_2024}. As their use becomes more widespread, there is a growing demand for advanced control strategies that ensure reliable operation in challenging, real-world conditions.

In this paper, we address the problem of tracking the position of a payload suspended from a UAV. Our focus lies in enabling reliable deployment in field conditions while minimizing hardware requirements. We propose a solution that operates using only standard UAV hardware—specifically, an RTK GNSS receiver and the onboard sensors available in most flight controllers (e.g., IMU, barometer, magnetometer...). Based on these inputs, we estimate and control the position of the suspended payload along a predefined reference trajectory.

\begin{figure}[htbp]
	\centering
	\includegraphics[width = \linewidth]{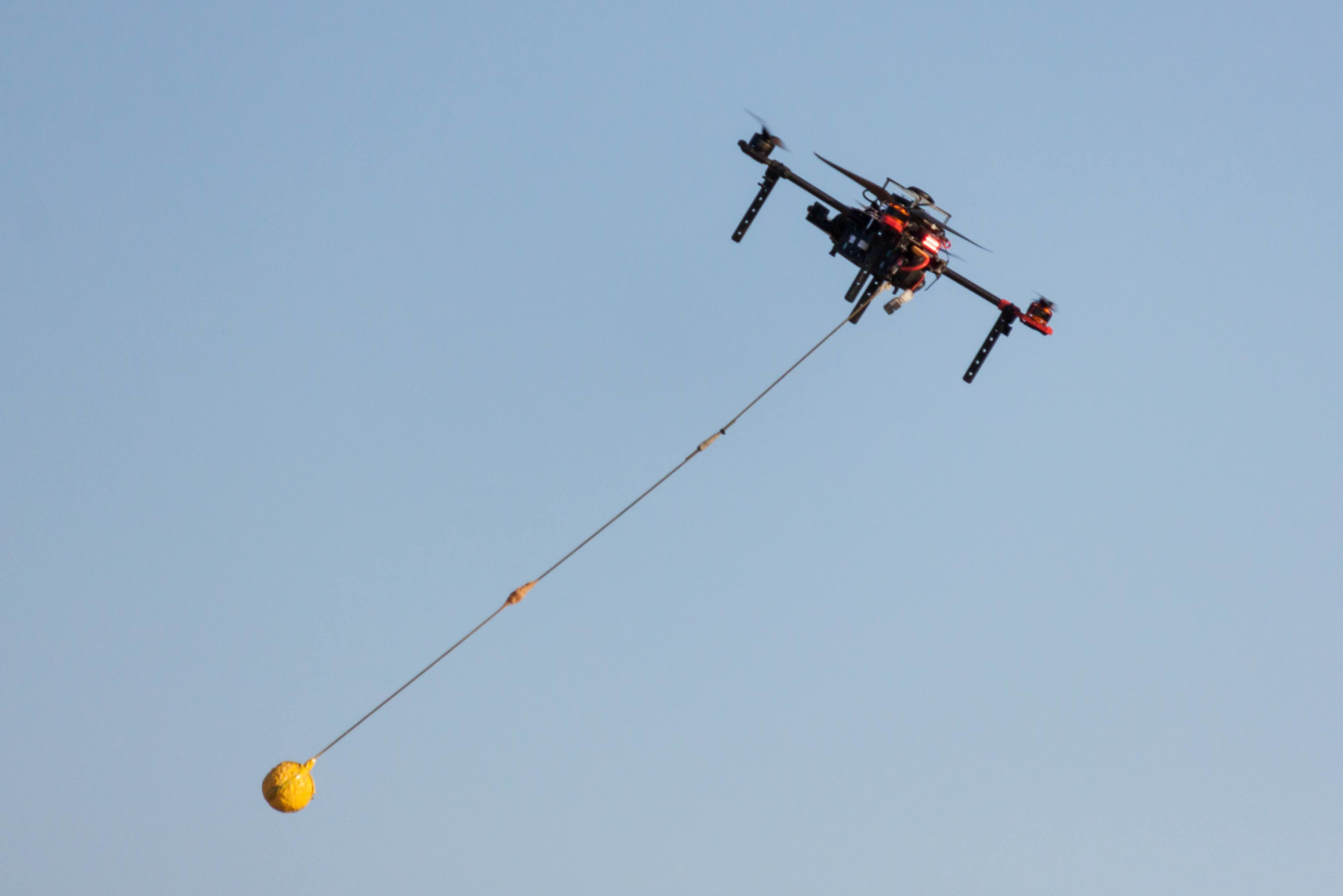}
	\caption{UAV carrying a cable suspended payload.}
	\label{fig:photo}
\end{figure}

In many practical applications, it is the position of the payload—rather than the UAV itself—that must be accurately controlled. For example, in agile pickup-and-delivery tasks, the end of the cable can reach the target before the UAV, exploiting the complex system dynamics. Similarly, in applications such as deploying sensors into hazardous or constrained environments, contacting surfaces with the payload, or precise mechanical interaction with the environment, controlling the payload’s position is of primary importance. Reducing the hardware and instrumentation requirements is key to simplifying deployment and lowering system cost—motivating our reliance solely on RTK GNSS and standard flight-controller sensors.

This problem presents significant challenges. The UAV-payload system is highly underactuated and exhibits complex, nonlinear coupled dynamics. Compounding the difficulty, the position and orientation of the payload are not directly measured in our setup. Existing state-of-the-art solutions either rely on adaptive controllers that treat the payload dynamics as disturbances \cite{li_autotrans_2023}, \cite{wang_impact-aware_2024}, or on perception-constrained control frameworks that use downward-facing cameras to track the payload \cite{li_pcmpc_2021}, \cite{recalde_es-hpc-mpc_2025}, \cite{sarvaiya_hpa-mpc_2025}. However, vision-based approaches introduce substantial complexity, cost, and sensitivity to environmental conditions, as well as constraints on maintaining payload visibility within the camera's field of view. A key limitation across many prior works is the lack of robust, fully onboard estimation tightly coupled with control.

We address this gap by designing and validating a lightweight, onboard-only state estimation and control framework capable of real-world payload tracking using generic UAV hardware. Our solution consists of a linear Kalman filter (LKF)-based estimator, an incremental model predictive controller (MPC), and a motion-planning component based on model predictive contouring control (MPCC). The incremental MPC formulation is inherently robust to zero-mean noise, which allows us to tune the estimator for low bias at the cost of tolerable noise. Meanwhile, the MPCC planner enables smooth transitions between reference waypoints and allows flexible trajectory shaping.

Experimental results show that the proposed system performs comparably to an idealized setup using ground-truth payload position, with only marginal performance degradation. The framework also demonstrates robustness to variations in payload parameters and has been successfully deployed in real-world outdoor environments, achieving results consistent with simulation. These findings highlight the viability of practical, low-cost UAV-based suspended payload tracking using only standard onboard sensors.

\section{\uppercase{Related Work}}
\label{sec:related_work}

Early research on UAVs with suspended payloads primarily focused on generating swing‑free trajectories and stabilizing hover under load. A pioneering contribution~\cite{palunko_trajectory_2012} demonstrated offline planning of swing‑free maneuvers using dynamic programming and input‑shaping techniques. Subsequent work~\cite{faust_learning_2013} applied reinforcement learning to synthesize swing‑minimized trajectories under uncertain dynamics. Later approaches, such as~\cite{sreenath_trajectory_2013}, leveraged differential flatness and geometric control to enable simultaneous tracking of both UAV and payload trajectories. The field has since expanded to include hybrid control frameworks~\cite{wang_impact-aware_2024}, disturbance‑robust planning~\cite{li_autotrans_2023}, and cooperative multi‑robot transportation strategies~\cite{zhang_if-based_2023}. However, accurate payload position tracking—particularly under onboard-only sensing—remains an open challenge.

Many works have focused exclusively on planning and control algorithms, typically evaluated in simulation or under motion-capture conditions. For instance,~\cite{li_autotrans_2023} proposed a real-time NMPC framework with whole-body safety guarantees, but evaluated it only in simulation or controlled laboratory environments. In~\cite{zhang_if-based_2023}, cooperative multi‑UAV planning using insetting‑formation methods was explored, though it relied heavily on external sensing and did not address onboard-only execution. Similarly,~\cite{tang_mixed_2015} presented MIQP‑based trajectory planning with hybrid dynamics and obstacle avoidance, assuming accurate state feedback. Overall, realistic onboard state estimation has often been abstracted or oversimplified in these control-centric studies.

A prominent line of research aims at estimating payload state—typically angles or position—to enable payload-aware control. Early methods employed inertial measurements or encoder-based observers to infer payload swing online, but required dedicated sensors~\cite{rego_suspended_2016}. More recent approaches have leveraged perception-driven estimation. For example, PCMPC~\cite{li_pcmpc_2021} proposed a perception-constrained MPC that fuses monocular camera and IMU data to estimate the cable direction and payload dynamics. Later, hybrid perception-aware frameworks such as HPA-MPC~\cite{sarvaiya_hpa-mpc_2025} and ES-HPC-MPC~\cite{recalde_es-hpc-mpc_2025} introduced advanced control and estimation strategies capable of handling slack-taut transitions, while enforcing visibility constraints via onboard cameras. Although these perception-enhanced MPC methods demonstrate high performance, they require reliable visual feedback and incur significant computational overhead. In contrast, fully onboard estimation using only standard UAV sensors—such as the flight controller IMU and RTK GNSS—is rarely addressed, leaving a gap in lightweight, deployable solutions for field-ready platforms.

To the best of our knowledge, no prior work has demonstrated robust payload position tracking and control in fully outdoor conditions using only GNSS (specifically RTK) and standard flight-controller sensors—without relying on motion capture, external cameras, or payload instrumentation. Our work addresses this gap by designing and validating a lightweight, onboard-only state estimation and MPC framework capable of tracking suspended payloads in real-world outdoor settings using standard UAV hardware. This enables practical deployment across a broad range of platforms and outdoor environments.

\section{\uppercase{Mathematical Model}}
\label{modelling}

The modeling of aerial systems transporting suspended payloads is a well-established problem in aerial robotics, with one of the most influential formulations introduced in \cite{palunko_trajectory_2012}. Their model, based on a Lagrangian derivation under the assumption of a taut, massless cable, captures the coupled motion between the unmanned aerial vehicle and the payload with sufficient fidelity for control design, while remaining analytically tractable. In this work, we adopt the core structure of Palunko’s model and extend it to account for aerodynamic damping effects on both the aerial vehicle and the payload. The inclusion of linear air-drag terms for each body provides a more realistic description of energy dissipation in outdoor environments, improving the model’s suitability for real-time state estimation and control under realistic conditions. This mid-level formulation maintains a balance between physical fidelity and computational efficiency, enabling onboard implementation without sacrificing robustness to unmodeled effects.

\subsection{Assumptions}
To ensure analytical tractability without significantly compromising realism, several simplifying assumptions are introduced. The cable linking the UAV and the payload is modeled as massless and always taut, with attachment points located precisely at the centers of mass (CoG) of both bodies. This eliminates any torque contributions or additional inertial effects due to the cable. The payload is treated as a rigid body of known mass $m_l$ and suspended at a fixed length $l$. Aerodynamic drag is included for both the UAV and payload, modeled as linear with respect to translational velocity. The payload is further assumed to exhibit symmetric aerodynamic characteristics, excluding rotational drag and shape-induced effects. Friction at the cable connection is neglected, thereby allowing the payload to rotate freely about the suspension point. Environmental disturbances such as wind are completely omitted to improve tracktability of the model. These modeling choices align with established literature \cite{palunko_trajectory_2012} and yield a manageable yet sufficiently expressive dynamic formulation.

\subsection{System Description}

\begin{figure}[htbp]
	\centering
	\includegraphics[scale = 1]{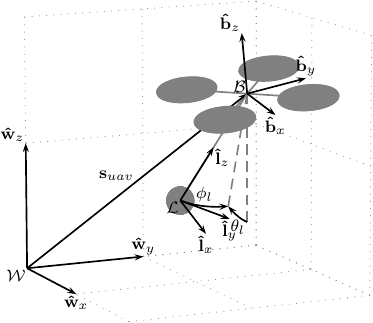}
	\caption{Coordinate systems and generalized coordinates of the UAV-payload system.}
	\label{fig:generalize_coordinates}
\end{figure}

The system configuration is described using three right-handed Cartesian coordinate frames, as illustrated in Fig.~\ref{fig:generalize_coordinates}. The world frame $\mathcal{W}$ is an inertial frame fixed to the Earth, with $\mathbf{\hat{w}}_z$ oriented upward. The body frame $\mathcal{B}$ is rigidly attached to the UAV, centered at its CoG, with $\mathbf{\hat{b}}_z$ aligned with the thrust direction. The load frame $\mathcal{L}$ is centered at the payload’s CoG, with $\mathbf{\hat{l}}_z$ aligned along the cable, and $\mathbf{\hat{l}}_x$, $\mathbf{\hat{l}}_y$ parallel to the corresponding body frame axes.

The UAV’s generalized coordinates in the world frame are defined by:
\begin{equation}
    \mathbf{q}_{uav} = \left[\begin{array}{cccccc}
        x & y & z & \theta & \phi & \psi
    \end{array}\right]^\intercal,
\end{equation}
where $\mathbf{s}_{uav} = \left[\begin{array}{ccc} x & y & z \end{array}\right]^\intercal$ denotes the UAV’s position, and $\Gamma = \left[\begin{array}{ccc} \theta & \phi & \psi \end{array}\right]^\intercal$ corresponds to its orientation, expressed via Tait-Bryan angles. The rotation from body to world frame is given by:
\begin{equation}
    \mathbf{R}_{\mathcal{B}\rightarrow\mathcal{W}}(\phi, \theta, \psi) = \mathbf{R}_z(\psi)\mathbf{R}_y(\theta)\mathbf{R}_x(\phi).
\end{equation}

Under the assumption of a taut cable, the payload’s relative position is parameterized by two angles:
\begin{equation}
    \mathbf{q}_{l} = \left[\begin{array}{cc}
        \theta_l & \phi_l
    \end{array}\right]^\intercal,
\end{equation}
which define the load-to-world rotation matrix as:
\begin{equation}
    \mathbf{R}_{\mathcal{L} \rightarrow \mathcal{W}}(\phi_l, \theta_l) = \mathbf{R}_x(\phi_l)\mathbf{R}_y(\theta_l).
\end{equation}
Using this formulation, the payload’s absolute position becomes:
\begin{equation}
    \mathbf{s}_l = \mathbf{R}_{\mathcal{L} \rightarrow \mathcal{W}} \left[\begin{array}{ccc}
        0 & 0 & -l
    \end{array}\right]^\intercal + \mathbf{s}_{uav}.
    \label{eq:model:load_position}
\end{equation}

\subsection{Equations of Motion}

Since the system's potential and kinetic energies do not depend on UAV attitude, the dynamics are formulated using the reduced generalized coordinate vector:
\begin{equation}
    \mathbf{q} = \left[\begin{array}{cc}
        \mathbf{s}_{uav} & \mathbf{q}_l 
    \end{array}\right]^\intercal.
\end{equation}
The Lagrangian is constructed in the standard form:
\begin{equation}
    \mathcal{L}\left(\mathbf{q}, \mathbf{\dot{q}}\right) = \mathcal{T}\left(\mathbf{\dot{q}}\right) - \mathcal{V}(\mathbf{q}),
    \label{eq:model:Lagrangian}
\end{equation}
where the potential energy is expressed as:
\begin{equation}
    \mathcal{V}(\mathbf{q}) = g \left[\begin{array}{ccc}
        0 & 0 & 1
    \end{array}\right]^\intercal \left(\mathrm{m}_{uav}\mathbf{s}_{uav} + \mathrm{m}_{l}\mathbf{s}_{l}\right),
    \label{eq:model:potential_energy}
\end{equation}
and the kinetic energy is given by:
\begin{equation}
    \mathcal{T}\left(\mathbf{\dot{q}}\right) = \frac{1}{2}\left(\mathrm{m}_{uav}\left|\left|\mathbf{\dot{s}}_{uav}\right|\right|^2 + \mathrm{m}_{l}\left|\left|\mathbf{\dot{s}}_{l}\right|\right|^2\right).
    \label{eq:model:kinetic_energy}
\end{equation}

The conservative force in the system originates from UAV thrust, modeled as:
\begin{equation}
    \mathbf{f}_{con}\left(\Gamma, F\right) = \left[\begin{array}{c}
         \mathbf{R}_{\mathcal{B}\rightarrow\mathcal{W}}(\Gamma)  \\
         \mathbf{0}
    \end{array}\right]
    \left[\begin{array}{c}
         0 \\
         0 \\
         F
    \end{array}\right],
    \label{eq:mode:conservative_force}
\end{equation}
where $F$ denotes the total collective thrust.

Aerodynamic drag represents the primary dissipative force and is incorporated as follows. The payload drag, projected onto angular coordinates, is modeled by:
\begin{equation}
	\mathbf{f}_{dis}^l(\dot{\mathbf{q}}, \mathbf{q}) = -\mathrm{d_{l}}\left[\begin{array}{ccc}
			1 & 0 & 0\\
			0 & -1 & 0
	\end{array}\right] \mathbf{R}_{\mathcal{L}\rightarrow\mathcal{W}}^{-1}(\theta_l, \phi_l) \dot{\mathbf{s}}_l^\mathcal{W},
	\label{eq:model:disLoad}
\end{equation}
while the radial drag component acts on the UAV as:
\begin{equation}
\begin{aligned}
	\mathbf{f}_{dis}^{uav}(\dot{\mathbf{q}}, \mathbf{q}) = -\mathrm{d}_{uav}\mathbf{\dot{s}}_{uav} \\
    -\mathrm{d_l} \mathbf{R}_{\mathcal{L}\rightarrow\mathcal{W}}(\theta_l, \phi_l) \left[\begin{array}{ccc}
			0 & 0 & 0\\
			0 & 0 & 0\\
			0 & 0 & 1
	\end{array}\right] \mathbf{R}_{\mathcal{L}\rightarrow\mathcal{W}}^{-1}(\theta_l, \phi_l) \dot{\mathbf{s}}_l^\mathcal{W}.
	\label{eq:model:disUAV}
\end{aligned}
\end{equation}
The complete expression for the dissipative force becomes:
\begin{equation}
    \mathbf{f}_{dis}\left(\mathbf{\dot{q}}, \mathbf{q}\right) =
    \left[\begin{array}{cc}
        \mathbf{f}_{dis}^{uav} & \mathbf{f}_{dis}^l
    \end{array}\right]^\intercal =
    -\mathbf{D}\left(\phi_l, \theta_l\right) \dot{\mathbf{q}}.
    \label{eq:mode:dissipative_force}
\end{equation}

Combining these elements and applying the Euler–Lagrange formalism yields the governing equations of motion:
\begin{equation}
\mathbf{M}(\mathbf{q})\ddot{{\mathbf{q}}} + \left(\mathbf{C}(\dot{\mathbf{q}}, \mathbf{q}) + \mathbf{D}(\mathbf{q})\right)\dot{{\mathbf{q}}} + \mathbf{g}(\mathbf{q}) = \mathbf{f}_{con}(\Gamma, F).
	\label{eq:model:equationsOfMotion}
\end{equation}

\subsection{Flight Controller Model}
The UAV's onboard flight controller (FCU) is abstracted as a set of decoupled first-order systems corresponding to each actuation channel. It tracks reference inputs $\mathbf{u}$ via internal state variables $\mathbf{x}_a = \left[\begin{array}{cccc}
    \theta & \phi & \psi & F
\end{array}\right]^\intercal$, governed by the following dynamics:
\begin{equation}
    \dot{\mathbf{x}}_a = \mathbf{A}_a \mathbf{x}_a + \mathbf{B}_a \mathbf{u},
    \label{eq:model:attitude}
\end{equation}
where $\mathbf{A}_a$ and $\mathbf{B}_a$ are diagonal matrices given by:
\begin{equation}
    \begin{aligned}
    \mathbf{A}_a &= \mathrm{diag}\left(-\frac{1}{\tau_1}, \dots , -\frac{1}{\tau_4}\right), \\
    \mathbf{B}_a &= \mathrm{diag}\left(\frac{\mathrm{K}_1}{\tau_1}, \dots , \frac{\mathrm{K}_4}{\tau_4}\right),
    \end{aligned}
\end{equation}
with $\tau_i$ and $\mathrm{K}_i$ denoting the time constants and gains, respectively, for each control channel.

\subsection{State Space Model}
The full nonlinear model, incorporating both physical dynamics and the flight controller, is expressed as:
\begin{equation}
\begin{aligned}
	\dot{\mathbf{q}} &= \nu,\\
        \dot{\nu}\!&=\!-\mathbf{M}^{-1}\!(\mathbf{q})\Big(\!\left(\mathbf{C}\!(\nu, \mathbf{q})\!+\!\mathbf{D}\!(\mathbf{q})\right)\nu\!+\!\mathbf{g}(\mathbf{q})\!-\!\mathbf{f}_{con}\!(\mathbf{x}_a)\Big), \\
	\dot{\mathbf{x}}_a &= \mathbf{A}_a \mathbf{x}_a + \mathbf{B}_a \mathbf{u}.
\end{aligned}
	\label{eq:model:nonlinStateSpace}
\end{equation}
It is noted that the mass matrix $\mathbf{M}$ becomes singular at $\theta_l = \frac{\pi}{2}$, imposing a constraint on the domain of explicit solutions.

For the purposes of control and estimation design, the system is linearized about the hover equilibrium with a motionless payload: $\theta_l = \phi_l = \SI{0}{\radian}$, $\dot{\theta}_l = \dot{\phi}_l = \SI{0}{\radian\per\second}$. The resulting linear time-invariant (LTI) model is:
\begin{equation}
    \dot{\mathbf{x}} = \mathbf{A}\mathbf{x} + \mathbf{B}\mathbf{u},
    \label{eq:model:linearSS}
\end{equation}
where the state vector is composed as:
\begin{equation}
    \mathbf{x} = \left[\begin{array}{ccc}
        \mathbf{q} &  \nu & \mathbf{x}_a
    \end{array}\right]^\intercal.
\end{equation}

\begin{figure*}[htbp]
	\centering
	\begin{adjustbox}{max totalsize={\textwidth}{\textheight}, center}
		 \tikzset{
  >=stealth,
  punkt/.style={
    rectangle,
    rounded corners,
    draw=black, very thick,
    text width=5em,
    minimum height=2em,
    text centered,
  },
  sum/.style = {draw,
	circle,
	node distance=1cm},
  arrow/.style={
    ->,
    very thick,
    shorten <=2pt,
    shorten >=2pt,
  }
}
\def\curvature{45}

\begin{tikzpicture}[node distance=2cm, auto,]

	\draw[blue!10, fill] (-13,-2.5) -- (-2.5,-2.5) -- (-2.5,1) -- (-13,1) -- (-13,-2.5);


	\node[punkt] (FCU) {Flight Controller};
	\node[punkt, left = of FCU, shift = {(-1.5, 0)}] (MPC) {MPC Controller};
	\node[punkt, left = of MPC, shift = {(-1.5, 0)}] (TG) {MPCC Planner};
        \node[punkt] (KF) at ($ (MPC)!.5!(FCU) + (-1.5,-1.5) $) {LKF Estimator};
        \node[punkt] (RTK) at ($ (FCU) + (0,-1.5) $) {RTK Estimator};

        \draw[->] ($(RTK.west)$) |- ($(KF.east)$) node [near start, shift = {(0.0, -0.5)}, text width=2.0cm] {\small{UAV's position and attitude}};
        \draw[->] ($(FCU.south)$) -- ($(RTK.north)$);
	\draw[->] ($(KF.west)$) -| ($(MPC.south)$) node [near start, shift = {(-1.0, 0.0em)}] {\small{state estimate}};
        \draw[->] ($(KF.west)$) -| ($(TG.south)$);
	\draw[->] ($(MPC.east)$) -- ($(FCU.west)$) node [midway, shift = {(0.0, 0.0em)}] {\small{attitude commands}};
        \draw[->] ($0.5*(MPC) - 0.5*(FCU) + (-1.5,0.0)$) -- ($(KF.north)$);
        \draw[->] ($0.5*(MPC) - 0.5*(FCU) + (1.0,0.0)$) |- ($(RTK.west) + (0,0.2)$);
	\draw[->] ($(TG.west) + (-2,0)$) -- ($(TG.west)$) node [near start, shift = {(0.3, 0.0em)}] {\small{target trajectory}};
        \draw[->] ($(TG.east) + (0,0.3)$) -- ($(MPC.west) + (0,0.3)$) node [midway, shift = {(0.0, 0.0em)}] {\small{optimized trajectory}};
	\draw[->] ($(MPC.west) + (0,-0.3)$) -- ($(TG.east) + (0,-0.3)$) node [midway, shift = {(0.0, 0.0em)}] {\small{predicted initial condition}};

\end{tikzpicture}
	\end{adjustbox}
	\caption{Architecture of the control framework. Blue block represents the proposed solution.}
	\label{fig:implementation:framework}
\end{figure*}
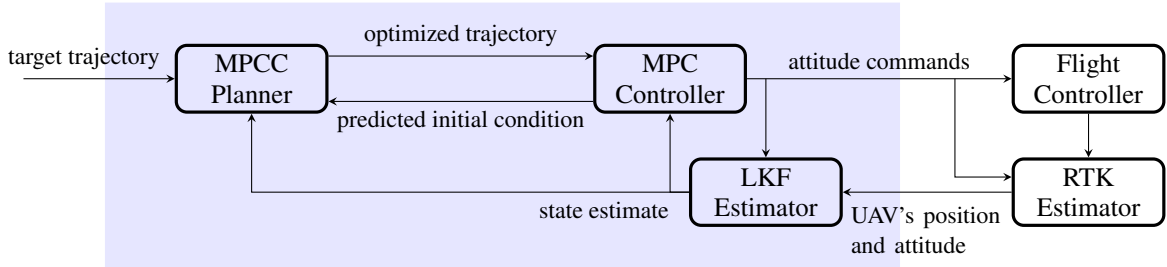

\section{\uppercase{Control Framework}}

The control framework consists of three components: a state estimator, a tracking controller, and a trajectory planner, as shown in Fig.~\ref{fig:implementation:framework}. The state estimator reconstructs the full system state, including unmeasured variables such as the payload position, using UAV position and attitude measurements in conjunction with the system model introduced in Section~\ref{modelling}. The tracking controller computes UAV attitude commands that ensure the payload tracks a desired trajectory by solving a constrained optimal control problem in a receding horizon fashion. Finally, the trajectory planner generates smooth, dynamically feasible reference trajectories online from sparse target states using temporally weighted optimization.

These components operate at different rates to ensure computational efficiency: the estimator and controller run at \SI{100}{Hz}, while the trajectory planner executes at \SI{1}{Hz}, enabling long-horizon planning without compromising real-time control performance.

\subsection{State Estimator}

We assume the UAV's position is measured via RTK GNSS, while its attitude is provided by the onboard flight controller. Rather than relying on raw measurements, we employ the RTK estimator from the MRS UAV System~\cite{baca_mrs_2021}, which provides built-in health checks and outlier rejection to ensure robust operation under field conditions.

To incorporate these measurements, the linear state-space model from equation~\eqref{eq:model:linearSS} is augmented with a measurement model:
\begin{equation}
    \mathbf{y} = \mathbf{C}\mathbf{x} =
    \begin{bmatrix}
        \mathbf{I}_3 & \mathbf{0}_{3\times7} & \mathbf{0}_{3\times3} & \mathbf{0}_{3\times1} \\
        \mathbf{0}_{3\times3} & \mathbf{0}_{3\times7} & \mathbf{I}_3 & \mathbf{0}_{3\times1}
    \end{bmatrix}
    \mathbf{x},
\end{equation}
where \( \mathbf{I}_3 \) is the \(3 \times 3\) identity matrix and \( \mathbf{0}_{a\times b} \) denotes an \(a \times b\) zero matrix.

The estimator is implemented as a Linear Kalman Filter (LKF)~\cite{kalman_1960} with empirically tuned noise covariances \( \mathbf{Q} \) and \( \mathbf{R} \), which balance smoothness and responsiveness (values listed in Table~\ref{tab:appendix_parameters}).

To account for variability in control loop timing—caused primarily by the onboard online optimization—the estimator employs forward Euler discretization with the current time step:
\begin{equation}
\bar{\mathbf{A}}(dt) = \mathbf{I} + dt\mathbf{A}, \quad
\bar{\mathbf{B}}(dt) = dt\mathbf{B},
\end{equation}
ensuring consistent and time-accurate state propagation even when the loop interval varies.

\subsection{Tracking Controller}

The tracking controller generates UAV attitude commands to track reference trajectories while accounting for constraints and dynamic limitations. It is formulated as a constrained optimal control problem and solved using Model Predictive Control (MPC) based on the linearized system dynamics.

The optimization problem is posed as a quadratic program (QP) with soft constraints:
\begin{align}
\min_{\substack{\tilde{\mathbf{x}}_t,\dots,\tilde{\mathbf{x}}_{t+N}\\ \tilde{\mathbf{u}}_t, \dots, \tilde{\mathbf{u}}_{t+N-1}\\ \mathbf{s}_t,\dots,\mathbf{s}_{t+N}}} \quad & 
\sum_{n=t}^{t+N} 
\begin{aligned}[t]
\frac{1}{2}
\begin{bmatrix}
\tilde{\mathbf{x}}_n \\
\tilde{\mathbf{u}}_n \\
1
\end{bmatrix}^\intercal
\begin{bmatrix}
\mathbf{Q}_n & \mathbf{0} & \mathbf{q}_n \\
\mathbf{0} & \mathbf{R}_n & \mathbf{0} \\
\mathbf{q}_n^\intercal & \mathbf{0} & 0
\end{bmatrix}
\begin{bmatrix}
\tilde{\mathbf{x}}_n \\
\tilde{\mathbf{u}}_n \\
1
\end{bmatrix}
+ {} \\
\frac{1}{2}
\begin{bmatrix}
\mathbf{s}_n^l \\
\mathbf{s}_n^u
\end{bmatrix}^\intercal
\begin{bmatrix}
\mathbf{Z}_n^l & 0 \\
0 & \mathbf{Z}_n^u
\end{bmatrix}
\begin{bmatrix}
\mathbf{s}_n^l \\
\mathbf{s}_n^u
\end{bmatrix},
\end{aligned}
\label{eq:controller:OCP:cost}
\end{align}

\begin{align}
\text{s.t.} \quad &
\tilde{\mathbf{x}}_{n+1} = \tilde{\mathbf{A}}_n \tilde{\mathbf{x}}_n + \tilde{\mathbf{B}}_n \tilde{\mathbf{u}}_n,& &n = t, \ldots, t+N-1,& \nonumber \\
&
\begin{bmatrix}
\underline{\tilde{\mathbf{u}}}_n \\
\underline{\tilde{\mathbf{x}}}_n
\end{bmatrix}
\leq
\begin{bmatrix}
\tilde{\mathbf{u}}_n \\
\tilde{\mathbf{x}}_n
\end{bmatrix}
+
\mathbf{s}_n^l,& &n = t, \ldots, t+N,& \nonumber \\
&
\begin{bmatrix}
\overline{\tilde{\mathbf{u}}}_n \\
\overline{\tilde{\mathbf{x}}}_n
\end{bmatrix}
\geq
\begin{bmatrix}
\tilde{\mathbf{u}}_n \\
\tilde{\mathbf{x}}_n
\end{bmatrix}
+
\mathbf{s}_n^u,& &n = t, \ldots, t+N.&
\label{eq:controller:OCP:contraints}
\end{align}

To gain finer control over the control actions and allow penalization of their rate, we adopt an incremental model formulation introduced in \cite{qin_survey_2003}:
\begin{equation}
    \begin{bmatrix}
         \mathbf{x}_{n+1}\\
         \mathbf{u}_{n+1}
    \end{bmatrix} =
    \begin{bmatrix}
         \mathbf{A} & \mathbf{B}\\
         \mathbf{0} & \mathbf{I}
    \end{bmatrix}
    \begin{bmatrix}
         \mathbf{x}_{n}\\
         \mathbf{u}_{n}
    \end{bmatrix} +
    \begin{bmatrix}
         \mathbf{B}\\
         \mathbf{I}
    \end{bmatrix} \Delta\mathbf{u}_n.
\end{equation}
This formulation helps prevent the propagation of noise from the state estimate to the control action and introduces robustness against estimation errors.

The payload position is estimated from the system state using a linearized form of Equation~\eqref{eq:model:load_position}, including control inputs:
\begin{equation}
    \tilde{\mathbf{y}}_n =
    \begin{bmatrix}
    \mathbf{s}_{\text{uav},n} +
    l
    \begin{bmatrix}
         -\theta_{l,n}\\
         \phi_{l,n}\\
         -1
    \end{bmatrix} \\
    \mathbf{u}_n
    \end{bmatrix}
    =
    \tilde{\mathbf{C}}
    \begin{bmatrix}
         \tilde{\mathbf{x}}_n \\
         1 
    \end{bmatrix}.
\end{equation}

Deviation from the reference is penalized through:
\begin{equation}
    \mathbf{Q}_n = \tilde{\mathbf{C}}^\intercal \, \mathrm{diag}(\mathbf{p}_{\mathbf{s}_l}, \mathbf{p}_{\mathbf{u}})\, \tilde{\mathbf{C}}, \quad
    \mathbf{R}_n = \mathrm{diag}(\mathbf{p}_{\Delta \mathbf{u}}),
\end{equation}
with values specified in Table~\ref{tab:appendix_parameters}.

Slack variables ensure feasibility under noise or disturbances, and the reference trajectory is introduced via a linear state term:
\begin{equation}
    \mathbf{q}_n = \left(\mathbf{x}_n^d\right)^\intercal \mathbf{Q}_n.
\end{equation}
The only states constrained in the optimization are translational velocity (per-axis) and the UAV’s attitude.
The optimization problem is solved by the HPIPM solver~\cite{frison_hpipm_2020} using partial condensation~\cite{axehill_controlling_2015}.

\subsection{Trajectory Planner}
\label{Planner}

The trajectory planner generates a dynamically feasible reference trajectory from target states using long-horizon optimization, complementing the fast controller operating at shorter time scales.

It solves an optimization problem structurally similar to the controller's formulation in \eqref{eq:controller:OCP:cost}, \eqref{eq:controller:OCP:contraints}, but adopts the Model Predictive Contouring Control (MPCC) paradigm introduced in \cite{romero_model_2022}. The key difference lies in the time-varying weighting matrix $\mathbf{Q}_i$ applied along the planning horizon:
\begin{equation}
    \mathbf{Q}_i = \omega_i \mathbf{Q},\quad i = 0,\dots,N,
\end{equation}
where
\begin{equation}
    \omega_i = \max_{k = 1,\dots,M} \exp\left(-\frac{\left(t_{r,k} - t_i\right)^2}{2\sigma^2}\right),\quad i = 0,\dots,N.
\end{equation}
Here, $t_{r,k}$ denotes the timestamp of the $k$-th state in the reference trajectory (consisting of $M$ states), and $t_i$ is the timestamp of the $i$-th step in the planning horizon. The parameter $\sigma$ controls the spread of the Gaussian kernels used for weighting.

This temporal weighting approach allows the planner to gradually transition between sparse reference points while maintaining smoothness and feasibility. For large values of $\sigma$, the weights vary slowly, encouraging adherence to the reference trajectory throughout the horizon. Conversely, small $\sigma$ values lead to sharp peaks in the weighting, guiding the system to closely match specific reference states while allowing flexibility in between.
The specific value of $\sigma$ used is provided in Table~\ref{tab:appendix_parameters}.

\begin{figure}[htbp]
	\centering
	\includegraphics[width = \linewidth]{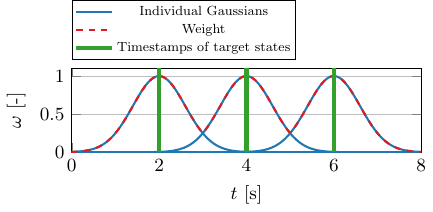}
	\caption{Illustration of the Gaussian weighting function $\omega_i$ along the planning horizon, centered at reference times $t_{r,k}$.}
	\label{fig:gaussian_weighting}
\end{figure}

A limitation of long-horizon MPC planners is that significant computation time may cause part of the computed trajectory to become outdated before it is applied. To mitigate this, the planner uses a predicted system state at time $t + t_{\text{plan}}$ as its initial condition, where $t_{\text{plan}}$ is an estimate of the planner’s execution time. This predicted state is available from the controller’s internal MPC prediction and ensures temporal alignment between planning and control.

\subsection{Limitations}
The proposed control framework assumes a constant UAV heading, as the payload-induced disturbances are modeled in the world coordinate frame and the system dynamics are linearized around a fixed heading. Another limitation is the assumption that the cable remains taut throughout the flight. Additionally, the framework requires prior knowledge of the payload’s parameters, such as mass and rope length. The current implementation is not adaptive and relies on these parameters being known and fixed. However, we consider these limitations to be acceptable in many practical scenarios. If necessary, they could be addressed through more complex adaptive, hybrid and nonlinear methods, albeit at the cost of increased computational demands.

\section{\uppercase{Verification}}

This section verifies the full control pipeline using the proposed metrics in both simulation and real-world experiments. While the system is capable of tracking densely sampled reference trajectories, such trajectories offer limited freedom for planning. In contrast, sparser trajectories allow for more planning optimizations and, when executed at higher speeds, better expose the dynamic properties of the suspended payload.

For instance, a slowly moving, densely sampled reference trajectory can be accurately followed using a conventional UAV position controller with a fixed vertical offset corresponding to the cable length. In such cases, the payload remains nearly stationary, and swing angles $\theta_l$ and $\phi_l$ are negligible. Consequently, the strengths and limitations of the proposed control framework are most apparent when tracking agile, sparse trajectories.

\subsection{Methodology}

Tracking errors on sparse reference trajectories are often difficult to interpret due to their potential inclusion of discontinuities or infeasible transitions. As such, absolute RMSE values can vary significantly depending on the chosen waypoint discretization.

\begin{figure}[htbp]
	\centering
    \includegraphics[width = \linewidth]{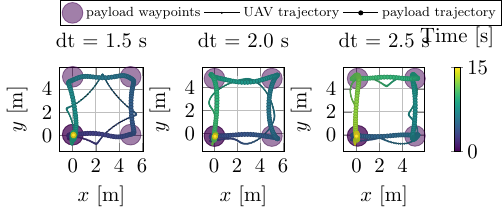}
	\caption{Open-loop planned trajectories for the square scenario, shown at different reference speeds.}
	\label{fig:square_trajectories}
\end{figure}

To establish a fair baseline, we introduce an \emph{open-loop (OL) planner}, which solves the same MPCC problem (refer to Section~\ref{Planner}) across the entire sparse reference to generate a dynamically feasible trajectory. These pre-planned OL trajectories, shown in Fig.~\ref{fig:square_trajectories}, serve as an idealized benchmark against which closed-loop performance can be measured.

We evaluate the full closed-loop system—including onboard estimation and MPC tracking—against this baseline using the \emph{relative RMSE degradation}:
\begin{equation}
  \Delta_{\mathrm{RMSE}} = 
  \frac{\mathrm{RMSE}(\mathbf{x}_\mathrm{exec}) - \mathrm{RMSE}(\mathbf{x}_\mathrm{OL})}
       {\mathrm{RMSE}(\mathbf{x}_\mathrm{OL})}
  \times 100\%,
\end{equation}
where $\mathrm{RMSE}(\mathbf{x})$ is computed over a trajectory $\mathbf{x}$ consisting of states $\mathbf{s}_0,\dots,\mathbf{s}_N$ as:
\[
  \mathrm{RMSE}(\mathbf{x}) =
  \sqrt{\frac{1}{N+1} \sum_{n=0}^{N} \left\| \mathbf{s}_n^{\mathrm{ref}} - \mathbf{s}_n \right\|^2},
\]
using zero-order hold to define a reference state $\mathbf{s}_n^{\mathrm{ref}}$ at step $n$.

The metric $\Delta_{\mathrm{RMSE}}$ captures how much tracking performance deteriorates compared to the ideal OL plan. Increased degradation indicates effects of estimation noise, limited prediction horizons, and model mismatch. We report $\Delta_{\mathrm{RMSE}}$ across varying trajectory speeds and payload parameters to assess robustness.

Two reference trajectories are used for evaluation: a square trajectory and a complex, custom-designed one. Both are composed of 3D waypoints with uniform temporal spacing defined by a time step $dt$.

The \emph{square trajectory} consists of alternating $5\,\mathrm{m}$ steps along the $x$ and $y$ axes, forming orthogonal segments to evaluate step response and speed sensitivity (Fig.~\ref{fig:square_trajectories}). The \emph{complex trajectory} includes changes in altitude, direction, and spacing, designed to challenge the controller with dynamically rich behavior (visualized in Fig.~\ref{fig:deployment:trajectory}).

\subsection{Simulation}

Simulations were performed using the high-fidelity Gazebo environment and the MRS UAV System, known for accurate sim-to-real transfer \cite{baca_mrs_2021}. The payload is modeled as a chain of 10 rigid links connected via spherical joints.
We analyze performance in terms of both trajectory speed and payload parameters.

\subsubsection{Effect of Trajectory Speed}

Simulations on the square trajectory were conducted at different speeds by varying the time step $dt \in \{1.5, 2.0, 2.5\}$\,s. Results in Table~\ref{tab:square:control} confirm that faster execution increases RMSE, as expected. For $dt = 2.5$\,s, the tracking error nearly matches that of the OL baseline, while for $dt = 1.5$\,s, the degradation remains below 40\%.

\begin{table}[htbp]
\vspace{-0.2cm}
\caption{Tracking RMSE degradation on square trajectory with varying speed ($\mathrm{l} = \SI{2}{\meter}$, $\mathrm{m}_l = \SI{1.5}{\kilo\gram}$).}
\label{tab:square:control}
\centering
\begin{tabular}{|c|c|c|c|}
  \hline
  & $\mathrm{RMSE}(\mathbf{x}_\mathrm{OL})$ & $\Delta_{\mathrm{RMSE}}^{\mathrm{gt}}$ & $\Delta_{\mathrm{RMSE}}$ \\
  $dt$ [s] & [m] & [\%] & [\%] \\
  \hline
  1.5 & 0.873 & 32.45 & 38.03 \\
  2.0 & 0.864 & 16.12 & 18.79 \\
  2.5 & 0.874 & 1.33 & 4.61 \\
  \hline
\end{tabular}
\end{table}

\begin{figure}[htbp]
	\centering
	\includegraphics[width = \linewidth]{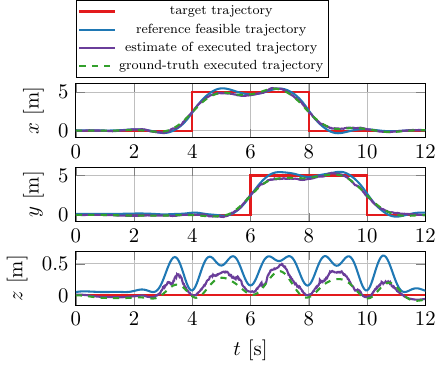}
	\caption{Closed-loop tracking on square trajectory ($dt = \SI{2}{\second}$).}
	\label{fig:square_control}
\end{figure}

Table~\ref{tab:square:estimation} shows estimation RMSEs for the swing angles and angular velocities. Estimation accuracy remains stable and correlates with trajectory excitation levels.
Minimal difference between using ground-truth and estimated states ($< 6\%$) demonstrates the estimator’s reliability. Figure~\ref{fig:square_control} illustrates a representative tracking result.

\begin{table}[htbp]
\vspace{-0.2cm}
\caption{Estimation RMSE for square trajectory with varying speed ($\mathrm{l} = \SI{2}{\meter}$, $\mathrm{m}_l = \SI{1.5}{\kilo\gram}$).}
\centering
\begin{tabular}{|c|c|c|c|c|}
  \hline
  & ${\theta}_l$ & ${\phi}_l$ & ${\dot{\theta}}_l$ & ${\dot{\phi}}_l$ \\
  $dt$ [s] & [rad] & [rad] & [rad/s] & [rad/s] \\
  \hline
  1.5 & 0.085 & 0.097 & 0.167 & 0.202 \\
  2.0 & 0.049 & 0.052 & 0.115 & 0.126 \\
  2.5 & 0.075 & 0.095 & 0.149 & 0.203 \\
  \hline
\end{tabular}
\label{tab:square:estimation}
\end{table}

\subsubsection{Effect of Payload Parameters}

To evaluate robustness against different payload configurations, we simulate the complex trajectory for varying cable lengths and masses (Table~\ref{tab:parameters:estimation}).

Estimation RMSE decreases with increasing mass, likely due to stronger excitation improving observability. Cable length has a minor effect, though longer cables show slightly increased errors.

\begin{table}[htbp]
\vspace{-0.2cm}
\caption{Estimation RMSE [rad] of payload angles across parameters ($dt = \SI{2}{\second}$).}
\centering
\begin{tabular}{c||c|c|c}
   & $\mathrm{l} = 1$\,m & $\mathrm{l} = 2$\,m & $\mathrm{l} = 3$\,m \\\hline\hline
   $\mathrm{m}_l = 0.5$\,kg & 0.201 & 0.145 & 0.186 \\
   $\mathrm{m}_l = 1.0$\,kg & 0.190 & 0.146 & 0.168 \\
   $\mathrm{m}_l = 1.5$\,kg & 0.179 & 0.141 & 0.163 \\
\end{tabular}
\label{tab:parameters:estimation}
\end{table}

Tracking degradation (Table~\ref{tab:parameters:control}) generally decreases with payload mass. However, results reveal that $\mathrm{l} = 2$\,m consistently leads to higher errors, suggesting that dynamic compatibility between the payload and trajectory may outweigh pure underactuation effects.

\begin{table}[htbp]
\vspace{-0.2cm}
\caption{Relative tracking RMSE degradation [\%] across parameters ($dt = \SI{2}{\second}$).}
\centering
\begin{tabular}{c||c|c|c}
   & $\mathrm{l} = 1$\,m & $\mathrm{l} = 2$\,m & $\mathrm{l} = 3$\,m \\\hline\hline
   $\mathrm{m}_l = 0.5$\,kg & 4.50 & 5.76 & 5.66 \\
   $\mathrm{m}_l = 1.0$\,kg & 2.29 & 4.97 & 3.37 \\
   $\mathrm{m}_l = 1.5$\,kg & 1.53 & 4.26 & 1.62 \\
\end{tabular}
\label{tab:parameters:control}
\end{table}

\subsection{Real-world Deployment}

\begin{figure}[!t]
    \centering
    \begin{minipage}{0.5\columnwidth}
        \centering
        \includegraphics[width=\linewidth]{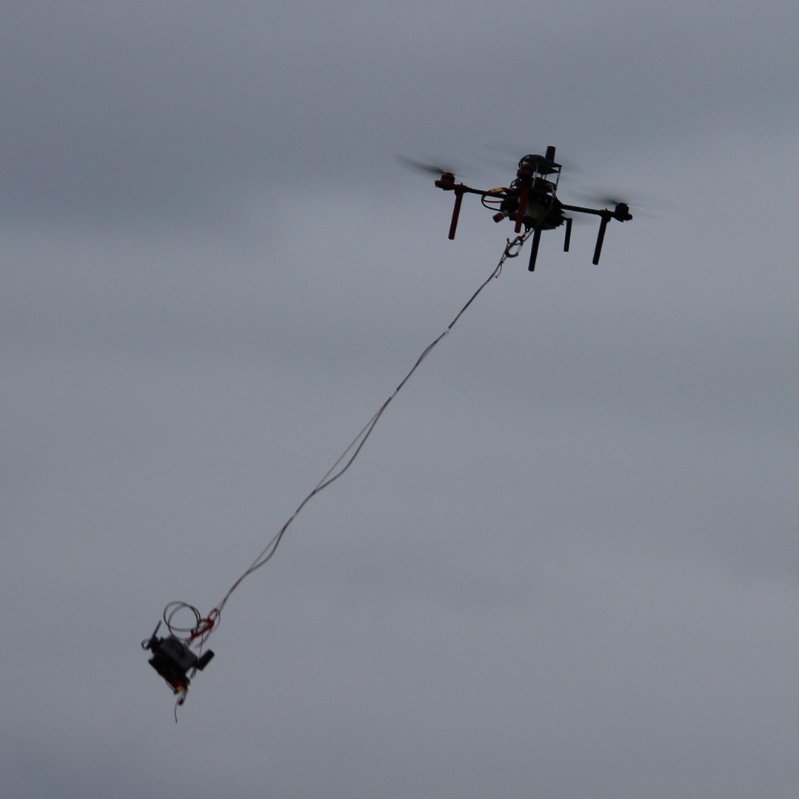}
        \caption*{(a) Photo from experiment}
    \end{minipage}\hfill
    \begin{minipage}{0.5\columnwidth}
        \centering
        \includegraphics[width=\linewidth]{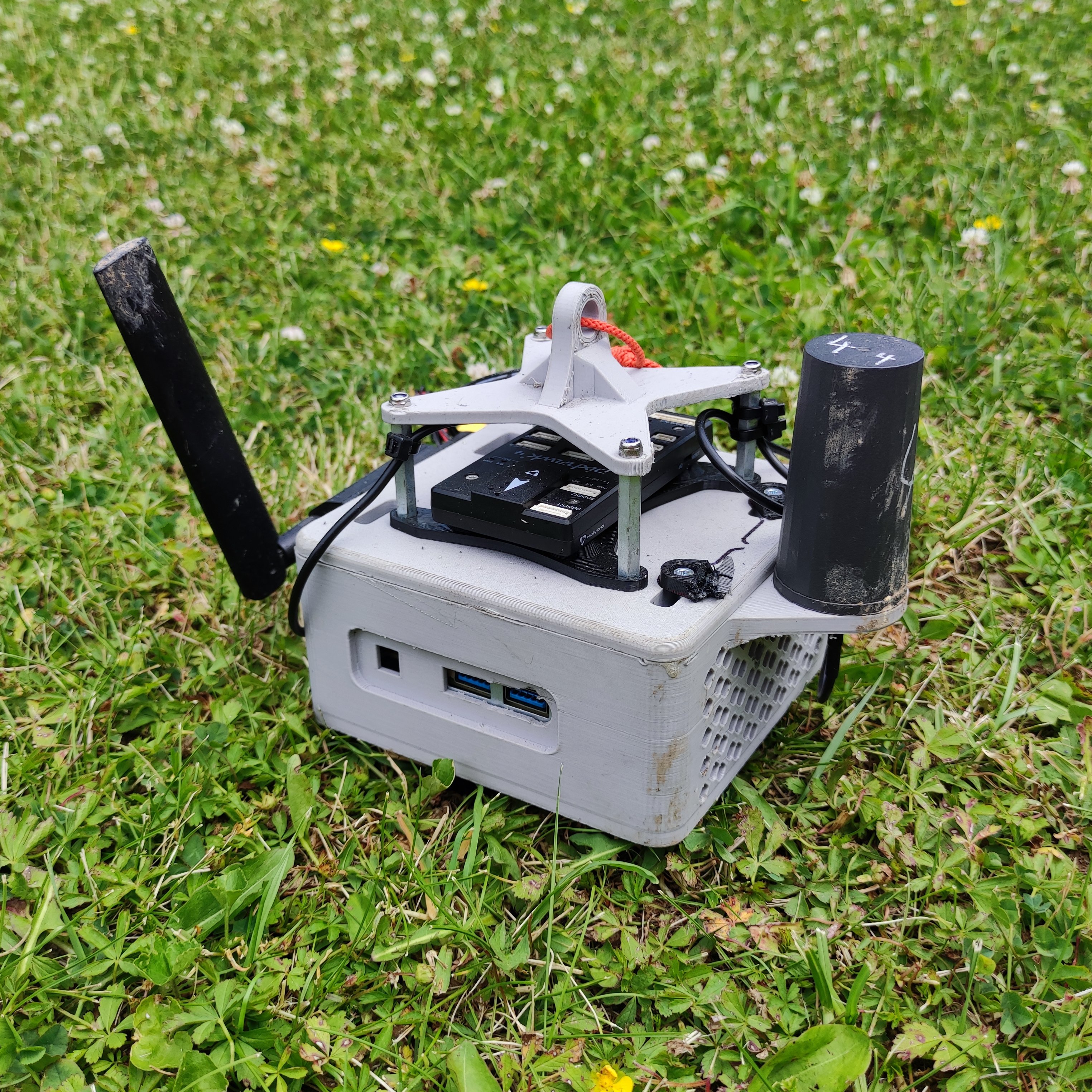}
        \caption*{(b) Payload with RTK GNSS}
    \end{minipage}
    \caption{Experimental hardware used in the field deployment.}
    \label{fig:deployment_photos}
\end{figure}

We deployed the proposed control pipeline on a Tarot T650 UAV platform equipped with an onboard Intel NUC computer, a Pixhawk 4 flight controller, and an Emlid Reach M2 RTK GNSS module to obtain global position measurements (Fig.~\ref{fig:deployment_photos}(a)).
A video of the real-world experiment is available at \href{https://mrs.fel.cvut.cz/papers/uav-with-cable-suspended-payload}{\url{https://mrs.fel.cvut.cz/papers/uav-with-cable-suspended-payload}}.
The payload consisted of a second Intel NUC, also equipped with an identical RTK GNSS receiver (Fig.~\ref{fig:deployment_photos}(b)), enabling direct ground-truth measurements of the payload’s 3D position.  
The payload mass was \SI{1.5}{\kilo\gram} and it was suspended using a \SI{2.0}{\meter} cable.
To ensure accurate data synchronization between the UAV and the payload computers, the two NUCs were physically connected via an Ethernet cable.

\begin{figure}[tbp]
	\centering
	\includegraphics[width = \linewidth]{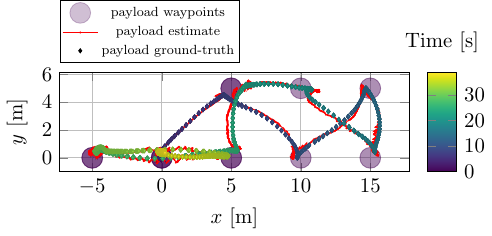}
	\caption{Executed trajectory during the real-world experiment.}
	\label{fig:deployment:trajectory}
\end{figure}
\begin{figure}[htbp]
	\centering
	\includegraphics[width = \linewidth]{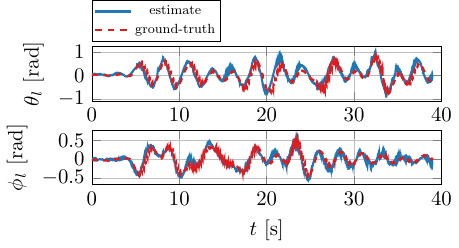}
	\caption{Comparison of estimated and ground-truth payload angles.}
	\label{fig:deployment:estimation}
\end{figure}

The payload was tasked with following a complex reference trajectory (see Fig. \ref{fig:deployment:trajectory}), discretized with $dt = \SI{3}{\second}$.
To assess the quality of the payload angle estimation, we compared the estimated angles $(\theta_l, \phi_l)$ with ground-truth angles reconstructed directly from the RTK GNSS position measurements.  
Fig.~\ref{fig:deployment:estimation} illustrates the comparison across the full trajectory, while Table~\ref{tab:deployment:estimation} summarizes estimation statistics for each angle.

\begin{table}[htbp]
\vspace{0.2cm}
\caption{Estimation accuracy on real-world trajectory}
\centering
\begin{tabular}{|c||c|c|}
  \hline
   & ${\theta}_l$ [rad] & ${\phi}_l$ [rad] \\
   \hline\hline
   RMSE & 0.178 & 0.078 \\
   STD  & 0.172 & 0.077 \\
   Bias & 0.046 & 0.012 \\
  \hline
\end{tabular}
\label{tab:deployment:estimation}
\end{table}

Estimation error is slightly higher in $\theta_l$, which corresponds to the direction of stronger excitation due to trajectory design.  
Importantly, both estimated angles exhibit very low bias relative to their standard deviation, indicating that the error is primarily due to zero-mean noise rather than modeling inaccuracies.  
This supports the validity of the dynamic model introduced in Section~\ref{modelling}.

\begin{figure}[htbp]
	\centering
	\includegraphics[width = \linewidth]{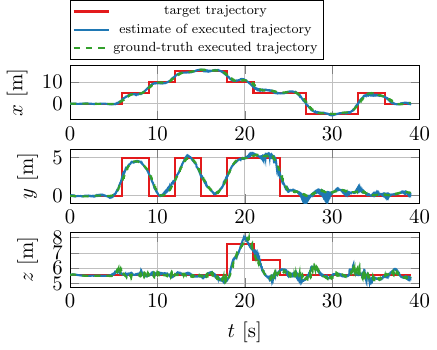}
	\caption{Trajectory tracking performance in real-world flight.}
	\label{fig:deployment:control}
\end{figure}

Using these state estimates, the UAV successfully tracked the desired trajectory.  
Fig.~\ref{fig:deployment:control} shows the tracking performance over the flight.  
Quantitative results are given in Table~\ref{tab:deployment:control}, which also presents simulation performance on the same trajectory to allow a sim-to-real comparison.

\begin{table}[ht]
\vspace{-0.2cm}
\caption{Real-world vs. simulation performance comparison}
\centering
\begin{tabular}{|c|c|c|c|}
  \hline
  Environment & \makecell{$\Delta_{\mathrm{RMSE}}$} & \makecell{Tracking\\RMSE\\{[m]}} & \makecell{Estimation\\RMSE\\{[rad]}}\\
  \hline \hline
  Simulation  & 1.36 & 1.563 & 0.134 \\
  Real-world  & 7.26 & 1.654 & 0.194 \\
  \hline
  \makecell{Sim-to-real\\gap} & \multicolumn{2}{c|}{+6\%} & +45\% \\
  \hline
\end{tabular}
\label{tab:deployment:control}
\end{table}

While estimation RMSE increases by 45\% in the real-world setting, tracking RMSE increases by only 6\%.  
This highlights the robustness of the MPC controller to noisy state estimates.  
The estimator's low bias ensures that tracking errors do not accumulate, while the controller’s incremental structure effectively filters out high-frequency estimation error.
These results demonstrate that the full control pipeline—from sparse trajectory planning to state estimation and control—remains performant under real-world conditions, validating the proposed framework as a practical solution for agile payload transport with suspended cables.

\section{\uppercase{Conclusions}}
\label{sec:conclusion}

We presented a complete control framework for real-time trajectory tracking of a suspended payload using an unmanned aerial vehicle equipped solely with standard onboard sensors—namely RTK GNSS and the flight controller's IMU. Unlike many prior approaches relying on external perception systems or additional payload instrumentation, our method operates in a fully onboard, field-deployable setup.

The proposed framework integrates a lightweight linear Kalman filter for state estimation, an incremental MPC for robust control under estimation noise, and a model predictive contouring control (MPCC) planner for smooth, adaptable trajectory generation. Our design leverages the robustness of incremental MPC to allow a bias-optimized estimator without sacrificing stability or tracking accuracy.

Experimental validation in simulation demonstrated that the proposed framework achieves performance close to ground-truth-based control, with tracking errors differing by only a few percent. The controller also showed strong robustness to variations in payload parameters. Furthermore, a field experiment confirmed the practical applicability of the system, achieving performance comparable to simulations and thereby validating the relevance of the simulated evaluations.

To our knowledge, this is the first demonstration of accurate and robust payload position tracking in outdoor environments using only common UAV hardware and onboard sensors. Our approach thus significantly lowers the barrier to deploying such systems in practical scenarios, paving the way for scalable applications in agile transport, sensor placement, and mechanical interaction tasks.

\section*{\uppercase{Acknowledgements}}
This work was supported by the European Union under the project *Robotics and Advanced Industrial Production* (reg. no. CZ.02.01.01/00/22\_008/0004590),  
by the Czech Science Foundation (GAČR) under research projects no. 23-07517S and no. 24-12360S, and by the CTU grant no. SGS23/177/OHK3/3T/13.

\bibliographystyle{apalike}
{\small
\bibliography{main}}

\begin{thebibliography}{}

\bibitem[Axehill, 2015]{axehill_controlling_2015}
Axehill, D. (2015).
\newblock Controlling the level of sparsity in {MPC}.
\newblock {\em Systems \& Control Letters}, 76:1--7.

\bibitem[Baca et~al., 2021]{baca_mrs_2021}
Baca, T., Petrlik, M., Vrba, M., Spurny, V., Penicka, R., Hert, D., and Saska,
  M. (2021).
\newblock The {MRS} {UAV} {System}: {Pushing} the {Frontiers} of {Reproducible}
  {Research}, {Real}-world {Deployment}, and {Education} with {Autonomous}
  {Unmanned} {Aerial} {Vehicles}.
\newblock {\em Journal of Intelligent \& Robotic Systems}, 102(1):26.

\bibitem[C. et~al., 2024]{c_path_2024}
C., M.~A., K., B., {Ramsundar}, and S., J. (2024).
\newblock Path {Planning} {Algorithm} for {UAV} {Based} {Water} {Quality}
  {Monitoring}.
\newblock In {\em 2024 3rd {Edition} of {IEEE} {Delhi} {Section} {Flagship}
  {Conference} ({DELCON})}, pages 1--5.

\bibitem[Colomina and Molina, 2014]{colomina_unmanned_2014}
Colomina, I. and Molina, P. (2014).
\newblock Unmanned aerial systems for photogrammetry and remote sensing: {A}
  review.
\newblock {\em ISPRS Journal of Photogrammetry and Remote Sensing}, 92:79--97.

\bibitem[Faust et~al., 2013]{faust_learning_2013}
Faust, A., Palunko, I., Cruz, P., Fierro, R., and Tapia, L. (2013).
\newblock Learning swing-free trajectories for {UAVs} with a suspended load.
\newblock In {\em 2013 {IEEE} {International} {Conference} on {Robotics} and
  {Automation}}, pages 4902--4909.
\newblock ISSN: 1050-4729.

\bibitem[Frison and Diehl, 2020]{frison_hpipm_2020}
Frison, G. and Diehl, M. (2020).
\newblock {HPIPM}: a high-performance quadratic programming framework for model
  predictive control⁎.
\newblock {\em IFAC-PapersOnLine}, 53(2):6563--6569.

\bibitem[Gode et~al., 2024]{gode_multi-stage_2024}
Gode, K., Gharat, K., Jogi, H., Sapkal, A., Thakar, R., Vishwakarma, S.,
  Talele, K., and Kulkarni, S. (2024).
\newblock Multi-{Stage} {UAV}-{Based} {System} for {Scalable} and {Accurate}
  {Crop} {Health} {Monitoring}.
\newblock In {\em 2024 {IEEE} {Space}, {Aerospace} and {Defence} {Conference}
  ({SPACE})}, pages 652--655.

\bibitem[Kalman, 1960]{kalman_1960}
Kalman, R.~E. (1960).
\newblock A {New} {Approach} to {Linear} {Filtering} and {Prediction}
  {Problems}.
\newblock {\em Journal of Basic Engineering}, 82(1):35--45.

\bibitem[Li et~al., 2021]{li_pcmpc_2021}
Li, G., Tunchez, A., and Loianno, G. (2021).
\newblock {PCMPC}: {Perception}-{Constrained} {Model} {Predictive} {Control}
  for {Quadrotors} with {Suspended} {Loads} using a {Single} {Camera} and
  {IMU}.
\newblock In {\em 2021 {IEEE} {International} {Conference} on {Robotics} and
  {Automation} ({ICRA})}, pages 2012--2018.
\newblock ISSN: 2577-087X.

\bibitem[Li et~al., 2023]{li_autotrans_2023}
Li, H., Wang, H., Feng, C., Gao, F., Zhou, B., and Shen, S. (2023).
\newblock {AutoTrans}: {A} {Complete} {Planning} and {Control} {Framework} for
  {Autonomous} {UAV} {Payload} {Transportation}.
\newblock {\em IEEE Robotics and Automation Letters}, 8(10):6859--6866.
\newblock Conference Name: IEEE Robotics and Automation Letters.

\bibitem[Murray and Chu, 2015]{murray_flying_2015}
Murray, C.~C. and Chu, A.~G. (2015).
\newblock The flying sidekick traveling salesman problem: {Optimization} of
  drone-assisted parcel delivery.
\newblock {\em Transportation Research Part C: Emerging Technologies},
  54:86--109.

\bibitem[Palunko et~al., 2012]{palunko_trajectory_2012}
Palunko, I., Fierro, R., and Cruz, P. (2012).
\newblock Trajectory generation for swing-free maneuvers of a quadrotor with
  suspended payload: {A} dynamic programming approach.
\newblock In {\em 2012 {IEEE} {International} {Conference} on {Robotics} and
  {Automation}}, pages 2691--2697.
\newblock ISSN: 1050-4729.

\bibitem[Qin and Badgwell, 2003]{qin_survey_2003}
Qin, S.~J. and Badgwell, T.~A. (2003).
\newblock A survey of industrial model predictive control technology.
\newblock {\em Control Engineering Practice}, 11(7):733--764.

\bibitem[Recalde et~al., 2025]{recalde_es-hpc-mpc_2025}
Recalde, L.~F., Sarvaiya, M., Loianno, G., and Li, G. (2025).
\newblock {ES}-{HPC}-{MPC}: {Exponentially} {Stable} {Hybrid} {Perception}
  {Constrained} {MPC} for {Quadrotor} with {Suspended} {Payloads}.
\newblock arXiv:2504.08841 [eess] version: 1.

\bibitem[Rego and Raffo, 2016]{rego_suspended_2016}
Rego, B.~S. and Raffo, G.~V. (2016).
\newblock Suspended load path tracking control based on zonotopic state
  estimation using a tilt-rotor {UAV}.
\newblock In {\em 2016 {IEEE} 19th {International} {Conference} on
  {Intelligent} {Transportation} {Systems} ({ITSC})}, pages 1445--1451.
\newblock ISSN: 2153-0017.

\bibitem[Romero et~al., 2022]{romero_model_2022}
Romero, A., Sun, S., Foehn, P., and Scaramuzza, D. (2022).
\newblock Model {Predictive} {Contouring} {Control} for {Time}-{Optimal}
  {Quadrotor} {Flight}.
\newblock {\em IEEE Transactions on Robotics}, 38(6):3340--3356.

\bibitem[Sarvaiya et~al., 2025]{sarvaiya_hpa-mpc_2025}
Sarvaiya, M., Li, G., and Loianno, G. (2025).
\newblock {HPA}-{MPC}: {Hybrid} {Perception}-{Aware} {Nonlinear} {Model}
  {Predictive} {Control} for {Quadrotors} {With} {Suspended} {Loads}.
\newblock {\em IEEE Robotics and Automation Letters}, 10(1):358--365.

\bibitem[Sikora et~al., 2023]{sikora_towards_2023}
Sikora, T., Markovic, L., and Bogdan, S. (2023).
\newblock Towards {Operating} {Wind} {Turbine} {Inspections} using a
  {LiDAR}-equipped {UAV}.
\newblock arXiv:2306.14637 [cs].

\bibitem[Sreenath et~al., 2013]{sreenath_trajectory_2013}
Sreenath, K., Michael, N., and Kumar, V. (2013).
\newblock Trajectory generation and control of a quadrotor with a
  cable-suspended load - {A} differentially-flat hybrid system.
\newblock In {\em 2013 {IEEE} {International} {Conference} on {Robotics} and
  {Automation}}, pages 4888--4895.
\newblock ISSN: 1050-4729.

\bibitem[Tang and Kumar, 2015]{tang_mixed_2015}
Tang, S. and Kumar, V. (2015).
\newblock Mixed {Integer} {Quadratic} {Program} trajectory generation for a
  quadrotor with a cable-suspended payload.
\newblock In {\em 2015 {IEEE} {International} {Conference} on {Robotics} and
  {Automation} ({ICRA})}, pages 2216--2222.
\newblock ISSN: 1050-4729.

\bibitem[Wang et~al., 2024]{wang_impact-aware_2024}
Wang, H., Li, H., Zhou, B., Gao, F., and Shen, S. (2024).
\newblock Impact-{Aware} {Planning} and {Control} for {Aerial} {Robots} {With}
  {Suspended} {Payloads}.
\newblock {\em IEEE Transactions on Robotics}, 40:2478--2497.

\bibitem[Zhang et~al., 2023]{zhang_if-based_2023}
Zhang, Y., Xu, J., Zhao, C., and Dong, J. (2023).
\newblock {IF}-{Based} {Trajectory} {Planning} and {Cooperative} {Control} for
  {Transportation} {System} of {Cable} {Suspended} {Payload} {With} {Multi}
  {UAVs}.
\newblock In {\em 2023 {IEEE}/{RSJ} {International} {Conference} on
  {Intelligent} {Robots} and {Systems} ({IROS})}, pages 635--642.
\newblock ISSN: 2153-0866.

\end{thebibliography}

\section*{\uppercase{Appendix}}
Table~\ref{tab:appendix_parameters} summarizes all key parameters used in the experiments, including those for the Linear Kalman Filter (LKF), the Model Predictive Controllers (MPC and MPCC), solver settings, and physical parameters of the UAV and payload.

\begin{table*}[ht]
\caption{Summary of parameters used in the framework. Parameters not specified for the MPCC planner are identical to those of the MPC controller.}
\centering
\renewcommand{\arraystretch}{1.2}
\setlength{\extrarowheight}{2pt}
\begin{tabular}{|p{7cm}|l|}
  \hline
  \textbf{Parameter} & \textbf{Value} \\
  \hline
  \rowcolor[gray]{0.9} \multicolumn{2}{|c|}{\textbf{Linear Kalman Filter}} \\
  Update rate & $\SI{100}{\hertz}$\\
  \makecell[l]{Process noise covariance $Q$ \\ $\left(\mathbf{s}_{uav}, \mathbf{q}_l, \mathbf{\dot{s}}_{uav}, \mathbf{\dot{q}}_l, \theta, \phi, F\right)$} & $\mathrm{diag}(1,1,1,30,30,100,100,100000,0.1,0.1,1,1,1)$ \\
  \makecell[l]{Measurement noise covariance $R$ \\ $\left(\mathbf{s}_{uav}, \theta, \phi\right)$} & $\mathrm{diag}(10,10,10,10,10)$ \\
  \hline

  \rowcolor[gray]{0.9} \multicolumn{2}{|c|}{\textbf{Incremental MPC Controller}} \\
  Update rate & $\SI{100}{\hertz}$\\
  Horizon length $N$ & 50 \\
  Sampling time $\Delta t$ & $\SI{0.05}{\second}$ \\
  \makecell[l]{State penalty matrix $Q$ parameters\\
  $\left(\mathbf{p}_{s_l}, \mathbf{p}_u\right)$
  } & $[10,10,10000,0,0,0.05]^\intercal$ \\
  \makecell[l]{Control penalty matrix $R$ \\ $\left(\Delta_{\theta_u}, \Delta_{\phi_u}, \Delta_{F_u}\right)$} & $\mathrm{diag}(100,100,5)$ \\
  Slack penalty matrix $R_{slack}$ & $\mathrm{diag}(10, \dots)$ \\
  Velocity bound & $\SI{10}{\meter\per\second}$\\
  Tilt control action bound & $\SI{0.75}{\radian}$\\
  \hline

  \rowcolor[gray]{0.9} \multicolumn{2}{|c|}{\textbf{MPCC Planner}} \\
  Replanning rate & 1 Hz \\
  Horizon length $N$ & 300 \\
  \makecell[l]{Control penalty matrix $R$ \\ $\left(\Delta_{\theta_u}, \Delta_{\phi_u}, \Delta_{F_u}\right)$} & $\mathrm{diag}(500,500,5)$ \\
  Tilt control action bound & $\SI{0.5}{\radian}$\\
  Contouring kernel variance $\sigma$ & 0.25 \\
  \hline

  \rowcolor[gray]{0.9} \multicolumn{2}{|c|}{\textbf{Solver}} \\
  Optimization solver & HPIPM \\
  Formulation & Partial condensing \\
  Sparsity level & 10 \\
  \hline

  \rowcolor[gray]{0.9} \multicolumn{2}{|c|}{\textbf{Model}} \\
  UAV mass $m_u$ & $\SI{3.5}{\kilo\gram}$ \\
  Payload air drag coefficient $d_l$ & 0.1 \\
  UAV air drag coefficient $d_{uav}$ & 0.1 \\
  \makecell[l]{Flight Controller Unit (FCU) model gains $K_i$ \\ $\left(\theta, \phi, F\right)$} & [1, 1, 1] \\
  \makecell[l]{FCU model time constants $\tau_i$ \\ $\left(\theta, \phi, F\right)$} & [0.2, 0.2, 0.05] \\
  \hline
\end{tabular}
\label{tab:appendix_parameters}
\end{table*}

\end{document}